\documentclass{article}

\usepackage{arxiv}

\usepackage[utf8]{inputenc} 
\usepackage[T1]{fontenc}    
\usepackage{hyperref}       
\usepackage{url}            
\usepackage{booktabs}       
\usepackage{amsfonts}       
\usepackage{nicefrac}       
\usepackage{microtype}      
\usepackage{lipsum}		
\usepackage{graphicx}
\usepackage{natbib}
\usepackage{doi}

\title{Evaluating ChatGPT as a Question Answering System: A Comprehensive Analysis and Comparison with Existing Models}

\author{ 
\href{https://bigdata.ui.ac.ir}{\includegraphics[scale=0.06]{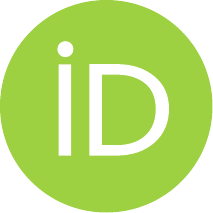}\hspace{1mm}Hossein Bahak, Farzaneh Taheri }\thanks{These two authors contributed equally to this work} \\
 Graduated Students of Computer Engineering,\\
	Big Data Research Group,\\
 University of Isfahan, Isfahan\\
	\texttt{hossein.bahak@eng.ui.ac.ir} \\\texttt{farzanehtaheri74@gmail.com} \\
  \And
    \href{https://bigdata.ui.ac.ir}{\includegraphics[scale=0.06]{orcid.pdf}\hspace{1mm}Zahra Zojaji } \\
    Assistant Professor of Computer Engineering,\\
	Big Data Research Group,\\
    Faculty of Computer Engineering,\\
    University of Isfahan, Isfahan\\
	\texttt{z.zojaji@eng.ui.ac.ir} \\
	\And
	\href{https://bigdata.ui.ac.ir}{\includegraphics[scale=0.06]{orcid.pdf}\hspace{1mm}Arefeh Kazemi } \\
	Big Data Research Group,\\
 Faculty of Computer Engineering,\\
 University of Isfahan, Isfahan\\
	\texttt{akazemi@fgn.ui.ac.ir} \\
}

\renewcommand{\shorttitle}{\textit{arXiv} Template}

\hypersetup{
pdftitle={A template for the arxiv style},
pdfsubject={q-bio.NC, q-bio.QM},
pdfauthor={David S.~Hippocampus, Elias D.~Striatum},
pdfkeywords={First keyword, Second keyword, More},
}
\begin{document}
\maketitle

\begin{abstract}
In the current era, a multitude of language models has emerged to cater to user inquiries. Notably, the GPT-3.5 Turbo language model has gained substantial attention as the underlying technology for ChatGPT. Leveraging extensive parameters, this model adeptly responds to a wide range of questions. However, due to its reliance on internal knowledge, the accuracy of responses may not be absolute. This article scrutinizes ChatGPT as a Question Answering System (QAS), comparing its performance to other existing QASs. The primary focus is on evaluating ChatGPT's proficiency in extracting responses from provided paragraphs, a core QAS capability. Additionally, performance comparisons are made in scenarios without a surrounding passage. Multiple experiments, exploring response hallucination and considering question complexity, were conducted on ChatGPT. Evaluation employed well-known Question Answering (QA) datasets, including SQuAD, NewsQA, and PersianQuAD, across English and Persian languages. Metrics such as F-score, exact match, and accuracy were employed in the assessment. The study reveals that, while ChatGPT demonstrates competence as a generative model, it is less effective in question answering compared to task-specific models. Providing context improves its performance, and prompt engineering enhances precision, particularly for questions lacking explicit answers in provided paragraphs. ChatGPT excels at simpler factual questions compared to "how" and "why" question types. The evaluation highlights occurrences of hallucinations, where ChatGPT provides responses to questions without available answers in the provided context.
\end{abstract}


\section{Introduction}
The ability to answer questions accurately and effectively is a critical component of natural language understanding and communication systems. Question answering (QA) tasks have gained significant attention in the field of Artificial Intelligence (AI) as they represent a fundamental challenge for developing intelligent systems that can comprehend and respond to user queries in a human-like manner. With the advent of large-scale language models, such as ChatGPT, which are based on advanced deep learning techniques, there is growing interest in evaluating their performance on QA tasks.

This paper aims to provide a comprehensive examination of ChatGPT's performance on the task of question answering. Despite the limited research in this area, no studies have evaluated ChatGPT's ability to extract answers from given paragraphs. When used as a QAS without any text for answer extraction, ChatGPT must rely on its knowledge to generate answers, which can be strongly influenced by training bias. Additionally, the problem of hallucination arises when ChatGPT is unable to clearly identify the answer to a specific question. This study focuses on investigating ChatGPT's ability to extract answers from given paragraphs and comparing its performance to cases where no surrounding passage is provided for the answer. As the ability of a QAS to extract answers from related passages is key to its performance, this concentration is justified for analyzing ChatGPT's capabilities as a QAS. 

We conducted a series of tests and evaluations using various question answering datasets to assess the effectiveness and accuracy of ChatGPT in providing precise answers to user queries. By leveraging well-established benchmark datasets, such as the Stanford Question Answering Dataset (SQuAD ), we can compare ChatGPT's performance against other state-of-the-art models and evaluate its capabilities in handling diverse question types and linguistic complexities. The choice of datasets is crucial for a comprehensive evaluation of ChatGPT's question answering performance. In addition to SQuAD , we utilized NewsQA as a widely recognized QA dataset for news domain and and PersianQuAD, which cover a broad range of topics and question types in Persian language. By including multiple datasets, we ensure a robust evaluation that encompasses different domains, question styles, and linguistic variations.

To measure the performance of ChatGPT on question answering, we employed well-established metrics such as accuracy, F1 score, and Exact Match (EM). These metrics allow us to assess the model's ability to provide correct answers and accurately capture the span of the answer within the given context. By quantifying ChatGPT's performance using multiple metrics, we can gain a comprehensive understanding of its strengths and weaknesses in tackling the complexities of question answering tasks. Furthermore, we analyzed and compared ChatGPT's performance across different question categories, including factoid and non-factoid questions. Factoid questions seek concise factual answers, while non-factoid questions require a deeper understanding of context and inference. By evaluating ChatGPT's performance on both types of questions, we can evaluate its ability to handle a wide range of question styles and information retrieval challenges. In addition to evaluating ChatGPT's performance, we also conducted an in-depth analysis of its responses to gain insights into the model's decision-making process. By examining both correct and incorrect answers, we aim to uncover patterns, limitations, and potential areas for improvement in ChatGPT's question answering capabilities. This analysis provides valuable insights into the model's reasoning abilities and aids in understanding its strengths and limitations. The findings of this study have significant implications for the development and enhancement of question answering systems. Understanding the performance and limitations of ChatGPT on question answering tasks contributes to advancing the field of natural language understanding and facilitates the development of more accurate and reliable AI systems. 

The key findings of the conducted research are summarized as follows:
\begin{itemize}
    \item The performance of ChatGPT was compared with that of other models pretrained in various tasks, including question answering. The results indicated that ChatGPT, as a general generative model, performs less effectively than task-specific models in answering questions.
    \item It was observed that ChatGPT's performance significantly improves when provided with context for answering questions, as opposed to when no surrounding paragraph is given.
    \item Experimental findings suggest that prompt engineering greatly enhances ChatGPT's precision, particularly when the answer to a question is not present in the provided paragraph. Specifically, a two-step prompt querying the presence of the answer prior to requesting it yields superior performance compared to a one-step query.
    \item ChatGPT demonstrates proficiency in answering simpler, factual questions compared to "how" and "why" question types.
    \item Evaluation of hallucinations revealed that ChatGPT provides responses to a significant percentage of questions for which no answer is available in the provided context.
  
\end{itemize}

The findings of this study have significant implications for the development and enhancement of question answering systems. Understanding the performance and limitations of ChatGPT on question answering tasks contributes to advancing the field of natural language understanding and facilitates the development of more accurate and reliable AI systems.

The remainder of this paper is organized as follows. In Section 2, we provide an overview of related work and highlight key advancements and challenges. Section 3 describes the evaluation framework employed for assessing ChatGPT's performance. Section 4 presents the results of the evaluation of the ChatGPT's responses from different points of view. Finally, we conclude the paper in Section 5, summarizing the key insights and contributions of our study.

\section{Related Work}
Since the research is related to both QAS and ChatGPT evaluation studies, we analyzed the related work into three sections including evaluation of ChatGPT, review of QASs and the evaluation of ChatGPT performance on QASs.
\subsection{Evaluation of ChatGPT}
There is limited research on the evaluation of ChatGPT as a QAS. However, a considerable research effort has been produced recently to evaluate various abilities of ChatGPT. Some general evaluation of ChatGPT was performed in several works (Bang
et al., 2023; Qin et al., 2023; Koco et al., 2023). Some researchers (Frieder et al., 2023; Borji, 2023) investigated its mathematical problem solving ability. Zhong et al. (2023) evaluated its understanding abilities. The performance of ChatGPT in fixing code bugs is examined by several works (Sobania et al., 2023; Surameery and Shakor, 2023; Haque and Li, 2023) and its overall applications in software engineering is studied in (Ma et. al, 2023). Again, it was also evaluated in translation (Jiao et al., 2023a; Jiao et al., 2023b; Gao et al. 2023b) and summarization tasks (Zhang et al. 2023b; Zhang et al. 2023c; Gao et al. 2023). Wang et al. (2023) studied the out-of-distribution behaviors of ChatGPT. Several studies (Borji, 2023; Bang et al., 2023; Zhang et al., 2023) analyzed the reasoning ability of ChatGPT. The success of ChatGPT for open information extraction was also investigated by Li et al. (2023). 

\subsection{ Question Answering Systems}

The field of Question Answering (QA) is widely recognized in Natural Language Processing (NLP), Information Extraction (IE), and Information Retrieval (IR). Initially, QA research emerged within the domain of IR. The first QA system, BaseBall \cite{green1961baseball}, was developed in 1961 to answer specific questions about American baseball games. This system utilized punched cards to process questions and retrieve answers from stored data on baseball games, employing linguistic patterns to understand question meaning. Another notable system, LUNAR \cite{woods1973progress}, was designed in 1973 to assist lunar geologists in finding answers related to NASA Apollo moon missions. This system leveraged information obtained from the missions to provide relevant answers. 

The Text Retrieval Conference (TREC) played a significant role in advancing QA research. In 1999, TREC introduced a QA track, offering researchers a dataset comprising news articles, questions, and corresponding answers \cite{voorhees2001trec}. TREC continued to organize QA tracks in subsequent years, leading to significant advancements in the QA task.

With the rise of the World Wide Web and the proliferation of web pages, researchers started utilizing web documents as the primary information source for QA systems \cite{zhu2021retrieving}. Modern QA systems are built upon web-based approaches, aiming to retrieve answers from online documents.

QA research has experienced notable progress in recent years, witnessing the development of numerous QA systems. Many of these systems, relying on deep learning techniques, require substantial amounts of training data. Notably, the majority of existing QA systems and datasets are predominantly available in English. The Stanford Question Answering Dataset (SQuAD ) \cite{rajpurkar2016SQuAD} is a well-known English QA dataset, consisting of approximately 100,000 questions based on Wikipedia articles. Other English QA datasets include MS Marco \cite{nguyen2016ms} and WikiQA \cite{yang2015wikiqa}, which were compiled by sampling questions searched on the Bing search engine, containing 100,000 and 3,000 samples, respectively. The Natural Questions dataset \cite{kwiatkowski2019natural} comprises 300,000 examples gathered from questions searched on the Google search engine. Additionally, the NewsQA dataset, sourced from CNN news webpages, contains over 100,000 instances and focuses on training QA systems for English news domains.

While numerous QA systems have been developed for non-English language, the scarcity of large-scale datasets poses a challenge. Consequently, the initial step in building a QA system for a non-English language involves creating a dataset specifically tailored to that language. Two main approaches are commonly employed: (1) translating existing English QA datasets such as SQuAD  using machine translation, and (2) constructing native QA datasets from scratch.

Some researchers have opted to translate SQuAD  into different languages and utilize the translated dataset to build QA systems. For instance, Carrino et al. (2020) translated SQuAD  into Spanish and developed a Spanish QA system. Mozannar et al. (2019) employed machine translation to convert 48,000 instances from SQuAD  into Arabic, subsequently building an Arabic QA system. Lee et al. (2018) performed machine translation of SQuAD  into Korean and then developed a Korean QA system. Croce et al. (2018) utilized semi-supervised translation techniques to create an Italian QA system by translating SQuAD  into Italian.

Alternatively, some researchers have focused on building native QA datasets for target languages and employed them to develop QA systems. Efimov et al. (2020) created a native QA dataset consisting of 50,000 instances for the Russian language, resulting in the development of a Russian QA system. Shao et al. (2018) developed a Chinese QA system using a native Chinese QA dataset comprising 30,000 instances. Similarly, Lim et al. (2019) and Keraronb et al. (2020) constructed QA systems for the Korean and French languages, respectively.

However, there are relatively few works dedicated to building QA systems for the Persian language. Veisi and Shandi (2020) developed a Persian medical QA system to address questions related to diseases and drugs. Abadani et al. (2021, 2021b) translated SQuAD  into Persian and built a general Persian domain QA system. Kazemi et al. (2022) created PersianQuAD, a large-scale Persian QA dataset comprising approximately 20,000 questions, and implemented a QA system for evaluation purposes. Lim et al. (2019) introduced a native answer selection dataset and a deep learning-based approach for Persian QA. Boreshban et al. (2018) constructed a religious QA dataset for answering questions in the Persian religious domain, utilizing it to develop a QA system.

\subsection{Evaluation of ChatGPT on QA systems}
In this section, we investigate the researches on evaluation of ChatGPT performance on question answering tasks. 
Omar et al. (2023)\cite{omar2023chatgpt} compared traditional question answering systems (i.e. KGQAN and EDGQA) with ChatGPT for knowledge graphs. Several experiments were conducted on four real-world knowledge graphs over different applications. The authors discussed the advantages and disadvantages of each approach and proposed future directions for improving knowledge graph chatbots. The authors prompted ChatGPT in three modes:  Default, Follow-up asking the entire list of response and Excel, requesting the answers in structured format. The QAS performance is evaluated in terms of determinism considering precision, recall and F1 score. The chatbot capabilities to produce correct and fluent answers is evaluated in terms of robustness, explainability and question understanding, through human manual assessment process. The paper suggests that ChatGPT provides more robust and explainable conversations compared to traditional QASs, but there are challenges to overcome, such as ensuring correctness, incorporating recent information and domain independency in responding. 
In a most recent study, Tan et al. (2023), developed an evaluation framework to analyze the reasoning capability of ChatGPT for answering complex questions. In the suggested framework, the performance of ChatGPT in question answering task is compared to several large language models like GPT-3 and GPT-3.5 upon 8 real world knowledge based datasets. To evaluate the correctness of ChatGPT answers, authors proposed a phrase-based answer matching strategy which matches the explanatory answer of ChatGPT with the specific gold answer of QA datasets. This procedure facilitates automatic evaluation of QA related metrics such as exact match. As a result, the paper utilized both automatic and human evaluation for investigating the performance of ChatGPT, which can be considered as an advantage over completely manual evaluation methods. This study found that, the performance of ChatGPT is overall better than other model in monolingual QA except for multi-hob, star-shaped reasoning as well as number and time based answers. Furthermore, it was revealed that ChatGPT outperforms other models on multilingual tests.  
In their recent study, Guo et al. (2023) present the Human ChatGPT Comparison Corpus (HC3), a dataset consisting of question-answer pairs in various domains such as finance, medicine, law, psychology, and open domain questions. The authors aim to compare the answers generated by ChatGPT with those provided by humans, using both Turing and Helpfulness tests. For the Turing test, two groups of individuals, including experts and amateurs, were asked to differentiate between answers generated by ChatGPT and those given by humans. The evaluation metric used in this test was the proportion of correctly identified ChatGPT-generated answers. In the Helpfulness test, human experts were requested to assess the helpfulness of ChatGPT answers compared to human answers for each question. The evaluation measure for the Helpfulness test was the proportion of instances where ChatGPT answers were deemed more helpful than human answers. The study conducted both qualitative and quantitative analyses on the results obtained from ChatGPT. The authors observed implicit differences between the responses generated by humans and ChatGPT. However, no comparison was made with existing question-answering systems. 
To sum up, in enumerated researches, there have been no examinations of ChatGPT's ability to extract answers from provided paragraphs. When utilized as a question-answering system (QAS) without any accompanying text for answer extraction, ChatGPT relies solely on its acquired knowledge to generate responses, which may be influenced by training biases. Moreover, the issue of hallucination arises when ChatGPT struggles to accurately identify the answer to a specific question. It seems that there is still an open issue on evaluating the ChatGPT performance in answer extraction from given paragraphs as a key challenge of QASs.

\section{The Evaluation Framework}

To assess the accuracy and efficiency of the responses obtained from the ChatGPT system, an evaluation framework has been employed (see Figure 1). This framework comprises three main components, including Prompt builder, ChatGPT and Answer evaluator. The overall structure of a QA dataset is in the form of a set of triples (P, Q, A), in which P refers to a textual paragraph, Q denotes a question about the paragraph and A stands for the corresponding answer. In some datasets, the answer to the question may not exist in the paragraph and an additional field determines that the answer does not exist in the corresponding paragraph. In the following, we refer to these two dataset types by type1 and type2 QA datasets, respectively. For each triple of the dataset, P\textsubscript{i} and Q\textsubscript{i} are entered into the Prompt builder section and a prompt is prepared and delivered to ChatGPT to ask Q given the P. Then the generated answer of the ChatGPT along with the gold answer A are passed to Answer evaluation component to analyze the performance using predefined evaluation metrics.

\begin{figure}
	    \centering
	    \includegraphics[width=0.9\linewidth]{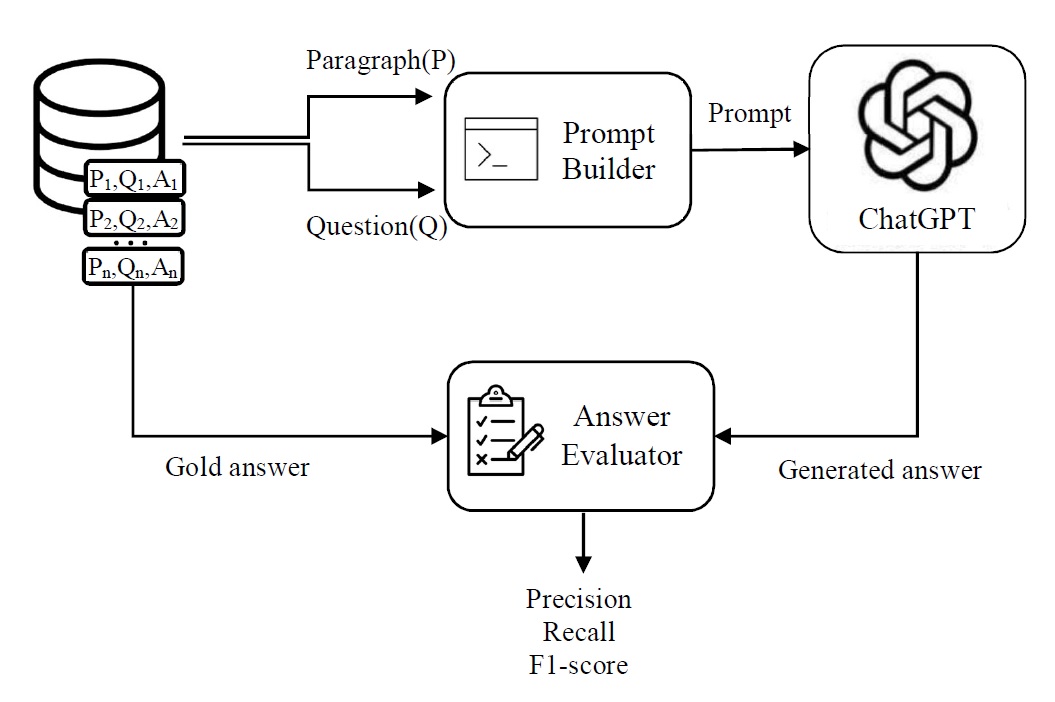}
	    \caption{The architecture of the evaluation framework}
	    \label{fig:enter-label}
	\end{figure}

\subsection{Prompt Builder}
The prompt builder is responsible for generating a proper prompt regarding the available data for feeding to ChatGPT. Therefore, two main prompts are designed for the type1 datasets, in which all the questions have their answers present in the given paragraphs and type 2 datasets which can also contain questions whose answers are not present in the given texts.

\begin{itemize}
    \item \textbf {Type 1 QA datasets: }
During the questioning process, an attempt has been made to only ask questions for which the answers exist in the text. The prompt designed for type 1 datasets is as follows:
\begin{verbatim}
"what is the answer of this question '" + Q + "' based on this paragraph '"
+ P +"'? say only the answer without saying the full sentence"
\end{verbatim}

    \item \textbf {Type 2 QA datasets: }
In the type 2 dataset, due to the presence of questions that do not have corresponding answers in the text, a two-step process is employed to obtain answers for the questions. In the first step, the text and question from the dataset are fed to the language model. the model is queried to determine whether the question exists in the provided text or not. The prompt for this step is as follows:
\begin{verbatim}
"Given the question '" + Q + "', determine if the following paragraph contains 
the corresponding answer: " + P + " say only 'Yes' or 'NO'"
\end{verbatim}

If the answer to the question is "Yes,", then we need to ask again to find the answer in the paragraph. The prompt for this second step is as follows:
\begin{verbatim}
"what is the answer of this question'" + Q + "'based on this paragraph'" + P +"'? 
say only the answer without saying the full sentence"
\end{verbatim}
It is important to note that only questions that are present in the text are used for evaluation. If a question does not exist in the text, nothing is asked from the model for that particular question during evaluation.
As we see in Table 1, NewsQA includes questions related to hallucinations. We give questions that we are sure have no answer to ChatGPT, and if answers are returned for those questions, it indicates a hallucination. If the text contains an answer to the asked question, it should return the answer; otherwise, it should return 'None.' In contrast, all SQuAD 1.1 questions can be found within the paragraphs provided. However, SQuAD  2.0 may include questions that are not present within the paragraphs, making it distinct in terms of question content.

\end{itemize}

\begin{table}[h]
\centering

\scriptsize 
\begin{tabular}{c|p{5cm}|c|p{4cm}|p{2cm}}
Dataset & Paragraph & Dataset Type & Prompt & GPT Response \\
\hline
NewsQA & NEW DELHI, India (CNN) -- 
A high court 
in northern India on Friday
acquitted a w... & Type 1 &what is the answer of this question "What is the number of children murdered based on this paragraph "A high court ... ? say only the answer without saying the full sentence & N/A (No specific answer provided)\\
\hline

SQuAD 2.0 & The Normans (Norman: Nourmands; French: Normands; Latin: Normanni) were the people who in the 10th and 11th centuries gave their name to Norma... & Type 2 &what is the answer of this question "In what country is Normandy located?" based on this paragraph "The Normans..."?
say only the answer without saying the full sentence & No\\

\hline
SQuAD  1.1 & Super Bowl 50 was an American football game to determine the champion of the National Football League (NFL) for the 2015 season. The American Football Conference (AFC) champion Denver Broncos defe ... & Type 1 & Which NFL team represented the NFC at Super Bowl 50? & Denver Broncos\\
\end{tabular}
\caption{Example of Prompt Builder}
\end{table}

As depicted in Table 1, the prompt builder encompasses two distinct types. This tailored approach ensures a more nuanced understanding of the data and contributes to the precision of generated responses, resulting in a better exact match.

\subsection{ChatGPT}
ChatGPT is one of OpenAI's important products with the capability of human-like language interaction. This product utilizes the GPT-3.5-turbo language model to provide responses to posed questions. In this research, an API connected to this language model has also been employed. Due to limitations on the number of requests and the number of tokens sent, significant challenges arose in obtaining information from the model. To overcome these challenges, various methods were used. For example, to address the issue of token limitations, prompt engineering and character limit considerations were used. To tackle the time constraint challenge, we resolved it by adjusting the request rate to 3 requests per minute (one request every 20 seconds).

\subsection{Answer Evaluator}
In order to evaluate the responses generated by ChatGPT, we initiate the process by normalizing both the ground truth sentence, which encapsulates the correct answer to the question within the dataset, and the response from ChatGPT. This normalization entails the removal of punctuation marks, special characters, and white spaces, along with converting all characters to lowercase. Following this preprocessing step, we proceed to compare these two sentences at the character level. We employ metrics such as Exact Match, F1 score, and Recall to quantitatively assess the level of correspondence between the generated response and the ground truth. These measures serve as valuable benchmarks for evaluating ChatGPT in comparison to previous models that have operated on similar datasets. This comprehensive evaluation methodology enables us to make informed comparisons and draw meaningful conclusions about the performance of ChatGPT in relation to its predecessors.

\section{Experimental Results}
In this section, we first introduce the datasets used for evaluation. Next, we introduce the evaluation metrics employed. Subsequently, we present the obtained results and compare them with the results of other models. 
Finally, the impact of prompt variations on the conclusions is examined.
For this experiment, we employed Google Colab for the collection and analysis of data.

\subsection{Data}
We aim to evaluate ChatGPT's performance in question-answering (QA) tasks, with a particular emphasis on its effectiveness in handling queries across diverse languages, specific domains and datasets including unanswerable questions. 

Numerous QA datasets are available for assessing question-answering systems in the English language. In our investigation, we chose to use three renowned datasets to assess ChatGPT's proficiency in English QA tasks. These datasets include the Stanford Question Answering Dataset (SQuAD ), encompassing both its initial version SQuAD  1.1 \cite{rajpurkar2016SQuAD} and SQuAD  2.0 \cite{rajpurkar2018know}, as well as NewsQA\cite{trischler2016newsqa}. For assessments related to the Persian language, we utilized the PersianQuAD dataset \cite{kazemi2022persianquad}. Each dataset presents unique characteristics and challenges, enabling a comprehensive evaluation of the model's capabilities. 

\begin{itemize}
    \item \textbf{SQuAD  1.1:} SQuAD  1.1 is widely recognized as one of the preeminent QA datasets in the English language. It boasts an extensive collection of over 100,000 questions, thoughtfully curated with contributions from crowdworkers who generated questions based on specific Wikipedia articles. Annotators were tasked with formulating queries based on provided article paragraphs and specifying corresponding answers.
    
    \item \textbf{SQuAD  2.0:} SQuAD  2.0, the second version of SQuAD  1.1, introduces an additional layer of complexity by incorporating over 50,000 deliberately unanswerable questions, intricately designed by crowdworkers to resemble answerable counterparts closely.
    
    \item \textbf{NewsQA:} NewsQA is created in three stages, with crowdworkers divided into three groups: questioners, answerers and validators. In the first stage, questioners see only the article highlights and headlines and formulate questions. In the second stage, answerers review the crowdsourced questions and the full article to determine the answers within the article. In the third stage, validators assess the article, the question, and a set of unique answers selected by the answerers. The validators then choose the best answer from the candidate set or reject all of them. the evaluation focuses solely on questions that have valid answers within the given text. the dataset's quality and accuracy, making it a valuable resource for assessing the language model's performance, particularly in detecting instances of hallucination.
    
    \item \textbf{PersianQuAD:} PersianQuAD represents a pioneering initiative and marks the inception of a comprehensive, large-scale native machine reading comprehension dataset designed for Persian language question-answering. This dataset comprises a curated collection of approximately 20,000 questions meticulously crafted by native annotators. To build PersianQuAD, annotators were presented with paragraphs from Persian Wikipedia articles, after which they were asked to create questions related to the content while also highlighting the answer within the paragraph text.
\end{itemize}

The thoughtful selection of these datasets, along with our rigorous evaluation criteria, forms a robust foundation for the comprehensive assessment of ChatGPT's performance across diverse linguistic contexts within the realm of QA tasks. This endeavour aims to provide valuable insights into ChatGPT's capabilities and limitations, contributing to a deeper understanding of its effectiveness.

\subsection{Metrics}
The two common evaluation metrics used for assessing QA systems are: "Exact Match" and "$F_1$ score"\cite{rajpurkar2016SQuAD}. The same metrics are employed in our research. In this study, we also use the Recall measure.
\begin{itemize}
    \item Exact Match:
This metric calculates the percentage of predicted answers that precisely match any of the correct candidate answers.

\item (Macro-averaged) $F_1$ score:
This metric determines the average overlap between the predicted answers and the correct candidate answers. To calculate the overlap, both the predicted and candidate answers are treated as bags of words, disregarding word order. For each question, the $F_1$ score is determined as the maximum $F_1$ score over all of its candidate answers. The "(Macro-averaged) $F_1$ score is the mean of the $F_1$ scores across all questions.

\item Recall:  Recall measures the QA system's ability to find all the relevant answers from a given set of documents or data sources. It is calculated as the ratio of the number of correctly retrieved relevant answers to the total number of relevant answers.

\end{itemize}

By utilizing these three evaluation criteria for each of the four datasets used in the evaluation, a comprehensive assessment of the Chat GPT system's performance is achieved. These metrics enable a quantitative evaluation of the model's accuracy, completeness, and overall effectiveness in answering questions.
All the results presented in the tables are expressed in percentages.
\subsection{Comparing ChatGPT with other Language Models on the Question Answering Task}

In this section, a comprehensive examination of various datasets has been undertaken, employing the GPT-3.5-Turbo language model as a question-answering system. The resulting insights stem from a meticulous comparison between the GPT-3.5-Turbo language model and other well-established language models. These models have undergone extensive testing and evaluation across diverse datasets, including SQuAD, NewsQA, and PersianQuad, serving as robust benchmarks for gauging the capabilities of the GPT-3.5-Turbo language model within a broader context.

To ensure a nuanced evaluation of the GPT-3.5-Turbo, we strategically selected popular question-answering models tailored for each dataset. For SQuAD 1.1, our chosen models include LUKE, XLNet, and SpanBERT. \\

In Table 2, the values of Exact Match and recall for these preceding question-answering models surpass those of GPT-3.5-Turbo. This discrepancy can be attributed to the specialized development of these models explicitly for question-answering tasks.

In Table 3, which focuses on SQuAD 2.0, we opted for Retro-Reader, Retro-Reader on ALBERT, and ALBERT. Notably, the values of Exact Match and recall for GPT-3.5 are lower than those of the other models, and they also fall short of the metrics observed in SQuAD 1.1. The rationale behind this disparity lies in the presence of unanswerable questions within SQuAD 2.0. In instances where the answer to a question is absent in the provided paragraph, GPT-3.5 generates responses, resulting in diminished Exact Match and recall values compared to SQuAD 1.1.

In Table 4, as observed, the dataset NewsQA comprises relatively lengthy stories, making the selection of relevant words challenging for each language model. This challenge is notably evident in the relevant terms and it may contribute significantly to the model's lower recall. The reduction in relevant words will inevitably result in shorter final model responses compared to the answers present in the dataset. Additionally, the inference of certain questions from various parts of the text could be another factor contributing to the lower exact match (EM) scores. Analyzing the results from the table, it becomes apparent that achieving high performance on NewsQA is a complex task, requiring models to navigate through lengthy narratives and effectively capture relevant information for accurate responses. This insight sheds light on potential avenues for improving the performance of language models in handling comprehension tasks on datasets with extended and context-rich content.

In Table 5, we shift our focus to evaluating the capabilities of GPT-3.5 Turbo in question-answering tasks for languages other than English. To accomplish this, we conduct a comparative analysis, pitting GPT-3.5 against models specifically designed for the complexities of Persian Quad. The findings reveal a significant performance gap, with GPT-3.5 lagging behind its counterparts. This disparity stems from the fact that the Persian Quad model is purpose-built, having undergone rigorous training on the Persian Quad dataset and, importantly, on the Persian language itself. The Persian Quad model's concentrated focus on both dataset and language nuances clearly grants it a competitive advantage over GPT-3.5 Turbo's more generalized language understanding in the domain of non-English languages.

\begin{table}
    \centering
    \small 
    \begin{tabular}{|l|c|c|}
        \hline
        Model Name & Exact Match & Recall \\
        \hline
        LUKE \cite{luke2020deep}& 90.2 & 95.4 \\
        \hline
        XLNet \cite{xlnet2019generalized}& 89.898 & 95.080\\
        \hline
        SpanBERT \cite{spanbert2019improving}& 88.8 & 94.6\\
        \hline
        ChatGpt & 44.4 & 87  \\
        \hline
    \end{tabular}
    \caption{Assessing the Effectiveness of Various Language Models on SQuAD 1.1.}
\end{table}

\begin{table}
    \centering
    \small 
    \begin{tabular}{|l|c|c|}
        \hline
        Model Name & Exact Match & Recall \\
        \hline
        Retro-Reader \cite{zhang2020retrospective}& 90.578 & 92.978 \\
        \hline
        Retro-Reader on ALBERT \cite{zhang2020retrospective}& 90.115 & 92.580 \\
        \hline
        ALBERT \cite{lan2020albert}& 89.731 & 92.215 \\
        \hline
        ChatGpt & 18.121 & 65\\
        \hline
    \end{tabular}
    \caption{Assessing the Effectiveness of Various Language Models on SQuAD  2.0}
\end{table}

\begin{table}
    \centering
    \small 
    \begin{tabular}{|l|c|c|}
        \hline
        Model Name & Exact Match & Recall \\
        \hline
        RAG-original \cite{rag_domain_adaptation}& 4.33 & 7.92 \\
        \hline
        BERT+ASGen \cite{back2021learning}& 54.7 & 64.5 \\
        \hline
        AMANDA \cite{kundu2020questionfocused}& 48.4 & 63.7  \\
        \hline
        DecaProp \cite{tay2020decaprop}{(Yi Tay et al. 2019)}& 53.1 & 66.3  \\
        \hline
        ChatGpt for Questions with
        all validator-confirmed answers & 41.07 & 46.70 \\
        \hline
    \end{tabular}
    \caption{Assessing the Effectiveness of Various Language Models on NewsQA}
\end{table}

\begin{table}
    \centering
    \small 
    \begin{tabular}{|l|c|c|}
        \hline
        Model Name & Exact Match & Recall \\
        \hline
        PersianQA \cite{kazemi2022persianquad}& 78.8 & 82.97 \\
        \hline
        ChatGpt & 41 & 55\\
        \hline
    \end{tabular}
    \caption{Results on PersianQuad}
\end{table}

\subsubsection{Exploring Diverse Input Information Volumes}

The SQuAD  datasets, particularly SQuAD  v1.0, holds a prominent position in the realm of question answering, serving as a benchmark for numerous researchers and their model development efforts. To evaluate ChatGPT's performance on these two datasets, two scenarios were considered: one where the paragraph is provided to the language model and another where it is not.

Table 6 illustartes the state that the paragraph is not given to the language model and only the language model is asked to answer the corresponding question in the dataset. 

In Table 7 depicts the state that the paragraph is also given to the language model in the corresponding prompt. A notable observation is the significant improvement in F1 score, Exact Match, and Precision when the language model is provided with the paragraph. This clearly demonstrates ChatGPT's enhanced ability to identify accurate answers when given access to the relevant context.

To assess ChatGPT's performance in non-English languages, the PersianQuAD dataset was selected. Similar to SQuAD  1.1, PersianQuAD ensures that answers to all questions reside within the corresponding paragraphs. The evaluation results are presented in Table 8, revealing a consistent decline in F1 score, Exact Match, and Precision compared to the SQuAD  datasets. This observation suggests that ChatGPT's performance on non-English languages is not as good as its performance on English language.
To evaluate ChatGPT's ability to handle hallucinations, the NewsQA dataset was employed. Table 8 presents the results of ChatGPT on the NewsQA dataset, revealing its performance in terms of F1 score, Exact Match, and Precision. 

In Table 9, given the volume of input data to the model and the constraints within the language model, a limited number of tokens related to the response have been returned. However, considering the higher precision compared to recall, it can be inferred that the returned tokens include those containing the correct answer. Therefore, it can be said that a relatively small number of relevant tokens have been returned, but these tokens are closely related to the answers present in the dataset. 

Following the domain specifications of NewsQA and PersianQuad datasets, questions were formulated within designated paragraphs. The evaluation metrics employed are tailored for assessing the model's performance within these datasets, aligning with their specific domains.

\begin{table}[h]
    \centering
    \begin{tabular}{|l|c|c|c|}
    \hline
    \textbf{Dataset} & \textbf{F1 score (\%)} & 
    \textbf{Recall (\%)} & \textbf{Precision (\%)} \\
    \hline
    SQuAD 1.1  & 1.92 & 1.98 & 2.05 \\
    \hline
    SQuAD  2.0   & 2.51 & 2.53 & 2.72 \\
    \hline
    \end{tabular}
    \caption{Evaluation Metrics for asked Questions of SQuAD  1.1, SQuAD  2.0 without paragraph}
\end{table}

\begin{table}[h]
    \centering
    \begin{tabular}{|l|c|c|c|}
    \hline
    \textbf{Dataset} & \textbf{F1 score (\%)} & \textbf{Recall (\%)} & \textbf{Precision (\%)} \\
    \hline
    SQuAD  1.1  & 67 & 87 & 62 \\
    \hline
    SQuAD  2.0  & 56 & 65 & 54 \\
    \hline
    \end{tabular} 
    \caption{Evaluation Metrics for asked questions with paragraph}
\end{table}

\begin{table}[h]
    \centering
    \begin{tabular}{|l|c|c|c|}
    \hline
    \textbf{Metric} & \textbf{F1 score (\%)} & \textbf{Recall (\%)} & \textbf{Precision (\%)} \\
    \hline
    PrsianQuAD  &55 &65 &53 \\
    \hline
    \end{tabular}
    \caption{Evaluation Metrics for Persian Quad Dataset}
\end{table}

\begin{table}[h]
    \centering
    \begin{tabular}{|l|c|c|c|}
    \hline
    \textbf{Metric} & \textbf{F1 score (\%)} & \textbf{Recall (\%)} & \textbf{Precision (\%)} \\
    \hline
    All Questions & 10.292 & 15.66 & 51.14 \\
    \hline
    \end{tabular}
    \caption{Evaluation Metrics for All Questions of NewsQA}
\end{table}

\subsubsection{Impact of Prompt Modification}

The quality of questions directly affects the quality of answers obtained from language models. Using appropriate prompts is the most crucial factor in improving the response quality of language models.

Several important points should be considered when asking questions to language models, such as using explicit, simple, and concise prompts. Initially, the questions asked were somewhat complex, but after prompt modification and multiple iterations of question refinement, more accurate answers were obtained from the language model.

Given that some questions in SQuAD  2.0 dataset have no answers, separating questions with answers in the text from those without answers is essential for a more precise evaluation of the model. In the first step, the language model was asked, that the answer to the question is present in the text or not. If yes, answer it. However, the obtained answers were not accurate, resulting in an undesirable F1 score. In the next step, questions were asked in two stages, and the evaluation metrics improved.

as we see in Table 10 The precision, F1 score, and recall for both steps are as follows:
\begin{table}[htbp]
    \centering
    \begin{tabular}{|l|c|c|c|}
        \hline
        \textbf{Method} & \textbf{Precision (\%)} & \textbf{F1 Score (\%)} & \textbf{Recall (\%)} \\
        \hline
        \textbf{two step prompt} & 27 & 37 & 88 \\
        \hline
        \textbf{single step prompt} & 62 & 67 & 87 \\ 
        \hline
    \end{tabular} 
    \caption{evaluation of question answering on SQuAD 2.0 for different prompts}
    \label{tab:evaluation_metrics}
\end{table}

As evident from the results in the Table 10, changing the prompt to more relevant words has resulted in extracting more answers from the text, accompanied by a higher F1 score.

Sample questions asked in the first and second steps of evaluation:

\textbf{Step 1}:
\begin{verbatim}
"Given the question " + question + " , determine if the following paragraph contains the
corresponding answer: " + paragraph + " say only 'Yes' or 'NO'"
\end{verbatim}

\textbf{Step 2}:
\begin{verbatim}
"What is the answer to this question " + question + " based on this paragraph "+
paragraph +"? Only provide the answer without full sentence."
\end{verbatim}

\subsubsection{Evaluation on Question Types (how-, what-, who-, when-, where- and listquestions)}
To comprehensively assess the performance of ChatGPT across diverse question types, we undertake a classification of inquiries within the SQuAD 1.1 and SQuAD  2.0 datasets. This classification is grounded in the initial word of each question, and subsequently, we compute the F1 score, precision, and recall metrics. Specifically, the questions are categorized under distinct headings such as 'what,' 'how,' 'when,' 'where,' 'who,' 'which,' and 'why.'
The resulting outcome presents the calculated precision, recall, and F1 score values for both the SQuAD 1.1 and SQuAD 2.0 datasets.

As evident from Figures 2 and 3, ChatGPT exhibits lower F1 score , Precision, and Recall scores for questions beginning with "Why" and "How" compared to other question types. This performance disparity stems from the inherent complexity of "Why" and "How" questions, which demand a deeper understanding of causal relationships and underlying mechanisms. In contrast, ChatGPT excels at answering simpler, factual questions, as reflected in its higher scores for those question types.

\begin{figure}
    \centering
    \includegraphics[width=0.5\linewidth]{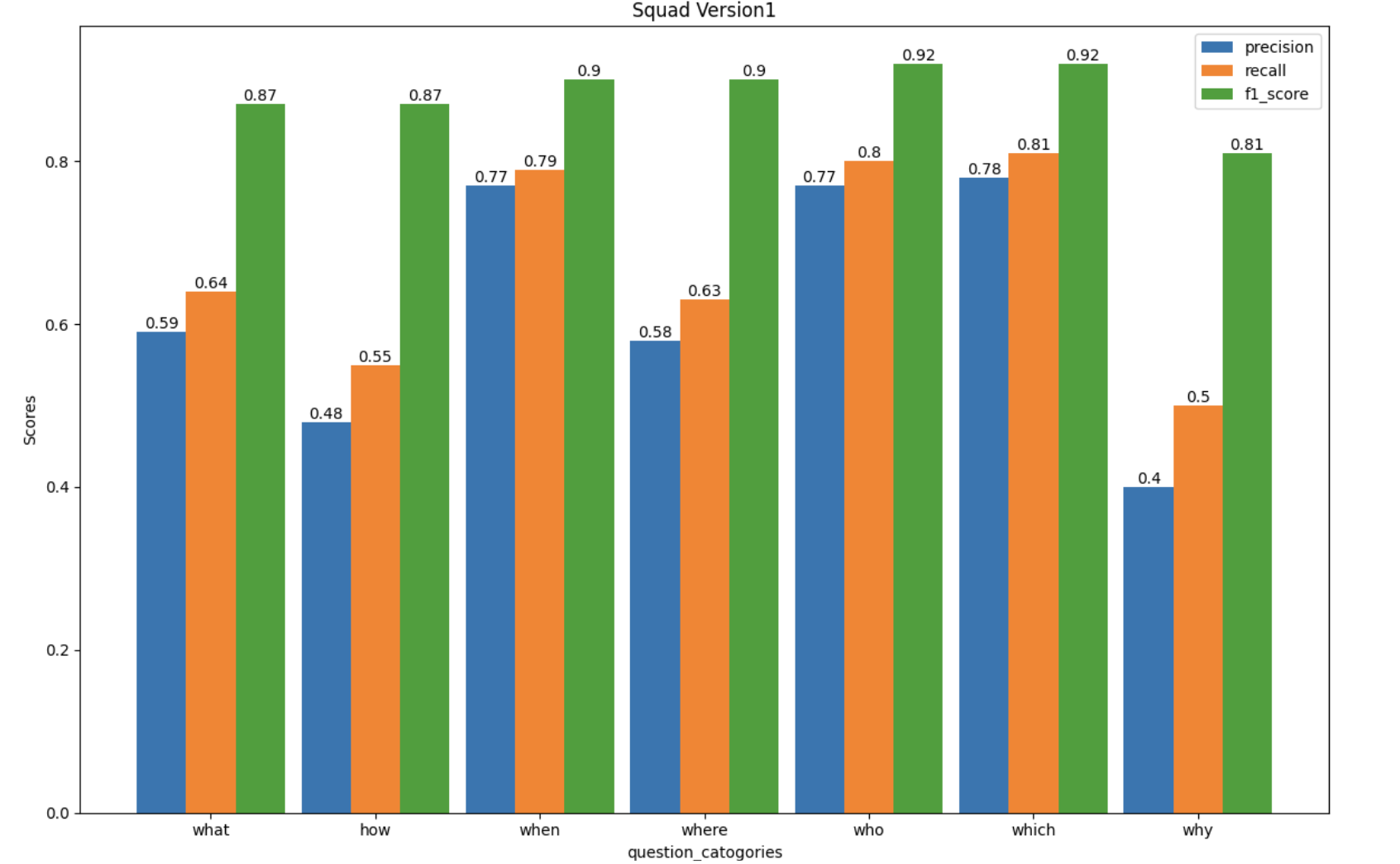}
    \caption{Histogram of computed evaluation metrics across various question types within the SQuAD 1.1 dataset}
    \label{fig:enter-label}
\end{figure}

\begin{figure}
    \centering
    \includegraphics[width=0.5\linewidth]{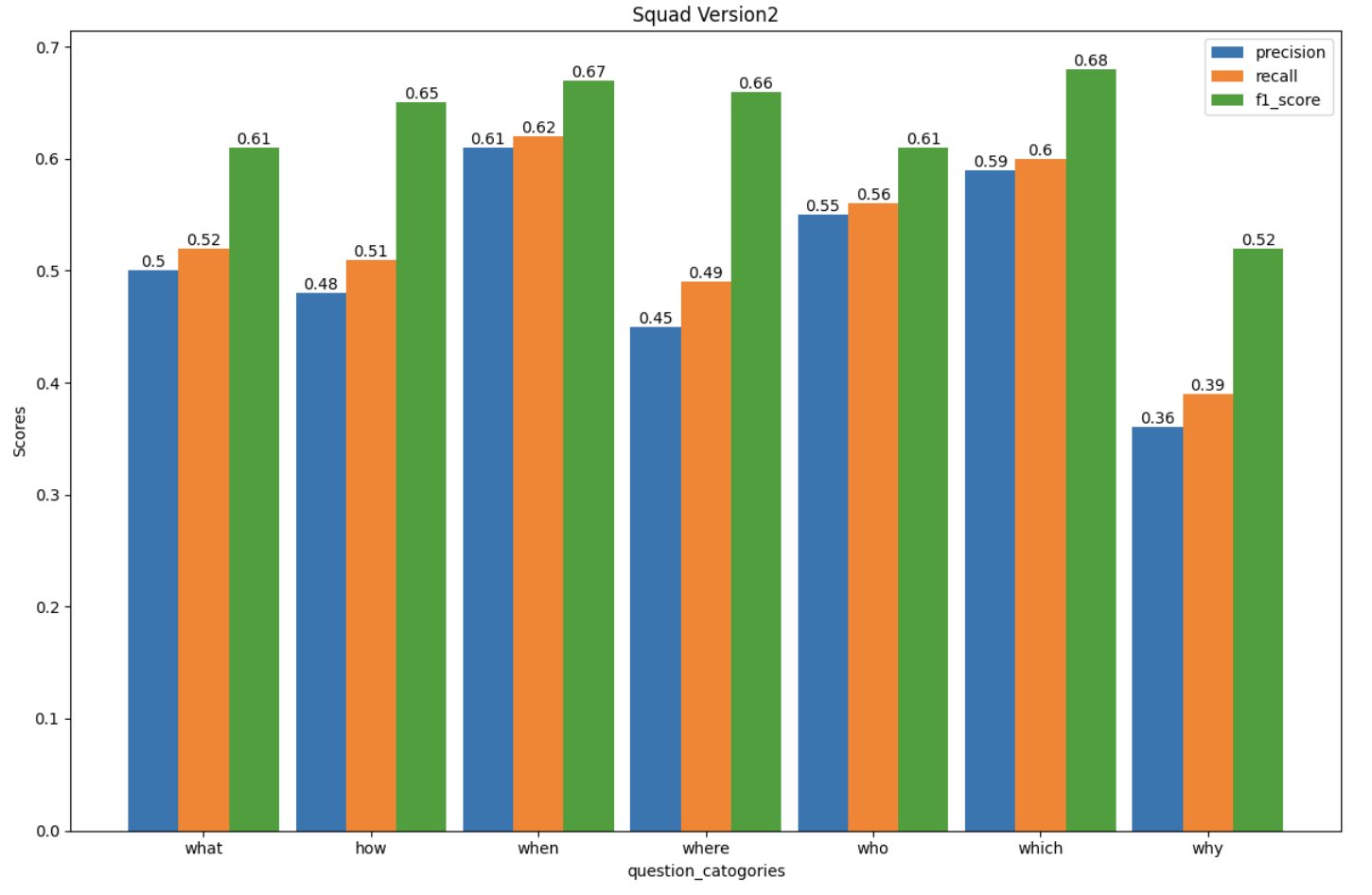}
    \caption{Histogram of computed evaluation metrics across various question types within the SQuAD 2.0 dataset}
    \label{fig:enter-label}
\end{figure}

\subsubsection{Evaluation on ChatGPT's hullicination}
It is important to highlight that certain datasets may contain questions that are not explicitly mentioned in the provided text. To address this challenge, a proposed evaluation framework is implemented in a manner that prevents the language model from generating responses in cases where the question is absent in the given text. The primary reason behind this approach is to avoid storing incorrect responses and their potential impact on the evaluation metrics.

Within this evaluation framework, an endeavor is made to examine the phenomenon of hallucination, which refers to instances where the language model generates responses that may seem plausible but lack factual basis. The process involves querying the language model about the existence of a given question in the provided text. Subsequently, when the model provides a response to a question that is confirmed to be absent in the text, it can be inferred that the response is a product of hallucination.

In the implemented code, two fields, isAnswerAbsent and isQuestionBad, are utilized to assess the presence or absence of an answer in the text. If these values are greater than zero, it signifies that some validators believe that the question's answer is not present in the given text or that the question has not been properly posed. If these values are equal to 1, it indicates that the question definitely does not exist in the given text, as all validators have voted against its presence. If a question has both isAnswerAbsent and isQuestionBad fields with values of 1, it means that the question is not present in the text. Consequently, if such a question is posed to the model, the response should be "NONE".

Prompt that used for this evaluation:

\begin{verbatim}
what is the answer to this question: {question_text}  in this text: {story_text} if there 
is no answer, just return NONE
\end{verbatim}

Further investigation into this matter will be undertaken based on the following results: (This dataset comprises approximately 580 thousand rows, but only 5598 rows have been utilized)

After formulating questions confirmed to lack definitive answers, we utilized the GPT-3.5-Turbo language model to generate responses. As presented in Table 11, the testing phase involved 5598 questions from the dataset. The results reveal that 91.31\% of the questions had at least one validator, implying the potential presence of incorrect answers. Despite the absence of a clear answer, the model endeavored to generate responses for these questions, drawing on the context provided by the validators.However, 8.68\% of the questions lacked any validators. The model still attempted to provide responses to these questions, suggesting that the generated answers could be characterized as hallucinations.

This evaluation process aims to identify and analyze instances of hallucination, ensuring that the generated responses are firmly grounded in the context of the given text and maintain factual accuracy. This critical assessment contributes to enhancing the reliability and trustworthiness of the Chat GPT system's performance, especially in the field of LLM.

\begin{table}[ht]
  \centering
  \begin{tabular}{|l|r|r|}
    \hline
    Question category & Count & Percent of all questions \\
    \hline
    At least one validator & 5112 & 91.31\% \\
    \hline
    Definitely have no answer & 486 & 8.68\% \\ 
    \hline
  \end{tabular}
\caption{Inquiries Posed and Corresponding Model-Generated Responses on NewsQA dataset}
\label{tab:results}
\end{table}

\subsubsection{Evaluation on Question Complexity}
To evaluate ChatGPT's performance on questions of varying difficulty, we calculated the F1 score for different ranges of question difficulty in the NewsQA, PersianQuad, SQuAD  1.1 and SQuAD  2.0 datasets.

In the realm of question-answering systems, the level of similarity between a question and its corresponding answer within a paragraph serves as a key determinant of system performance. Intuitively, when there is less similarity between the question and the answer, the task becomes more challenging for the QA system. This observation points to an inverse relationship between the difficulty of the task and the degree of similarity between the question and answer. To quantify this similarity, the Jacard Coefficient emerges as a valuable metric. 

Formally, the Jaccard Coefficient, denoted by J(Q, A), is calculated as follows:
\[ J(Q, A) = \frac{|Q \cap A|}{|Q \cup A|} \]

As outlined in (1), the Jacard Coefficient calculates similarity by considering the ratio of common words shared between the question and answer sentences to the total number of words in both the question and answer. With the Jacard Coefficient ranging between 0 and 1, a value of 0 signifies no lexical similarity between the question and the sentence, while a value of 1 indicates that the sentence encompasses all the words present in the question.

To assess ChatGPT's performance on questions of varying difficulty, we calculated the F1 score for different ranges of Jaccard similarity values across the NewsQA, PersianQuad, SQuAD 1.1, and SQuAD 2.0 datasets. Figures 4-7 depict the F1 scores corresponding to different levels of similarity between questions and their respective answers in the aforementioned datasets.

Figure 4 unveils a discernible pattern in ChatGPT's performance as it navigates through various Jacard similarity levels. In the context of SQuAD 1, ChatGPT reaches its zenith F1 score, registering an impressive 0.83, precisely when confronted with questions possessing a similarity level of 0.5. This intriguing observation implies that ChatGPT exhibits a commendable balance between precision and intricacy, showcasing its proficiency in addressing questions of moderate complexity.

Figure 5 intriguingly reveals that ChatGPT's F1 score in SQuAD 2 reaches its apex at 0.78 for questions with a Jaccard similarity level of 0.3. This observation suggests that ChatGPT's prowess is particularly pronounced when dealing with questions exhibiting lower similarities

In Figure 6, within the Persian Quad dataset, ChatGPT achieves a peak F1 score of 0.78, specifically observed for questions with a Jacard similarity of 0.09.

In Figure 7, considering the obtained numerical range of 120 for the difficulty chart of questions in the NewQA dataset, we opted to enhance the ease of interpretation and better comprehension of the figure by transforming the difficulty values into twenty-point intervals. Initially, when the average value of each interval was calculated and displayed, the overall figure appeared uniform. However, after conducting sampling within each interval and incorporating the obtained values into the figure, we observed that F1 score exhibits periodic behavior with increasing question difficulty. Interestingly, the level of difficulty seems to have a cyclic impact on F1 score, showing no direct correlation with increases or decreases in difficulty. These findings shed light on the nuanced relationship between question difficulty and the cyclic behavior of F1 score, providing valuable insights into the model's performance on different difficulty levels.

\begin{figure}
    \centering
    \includegraphics[width=0.5\linewidth]{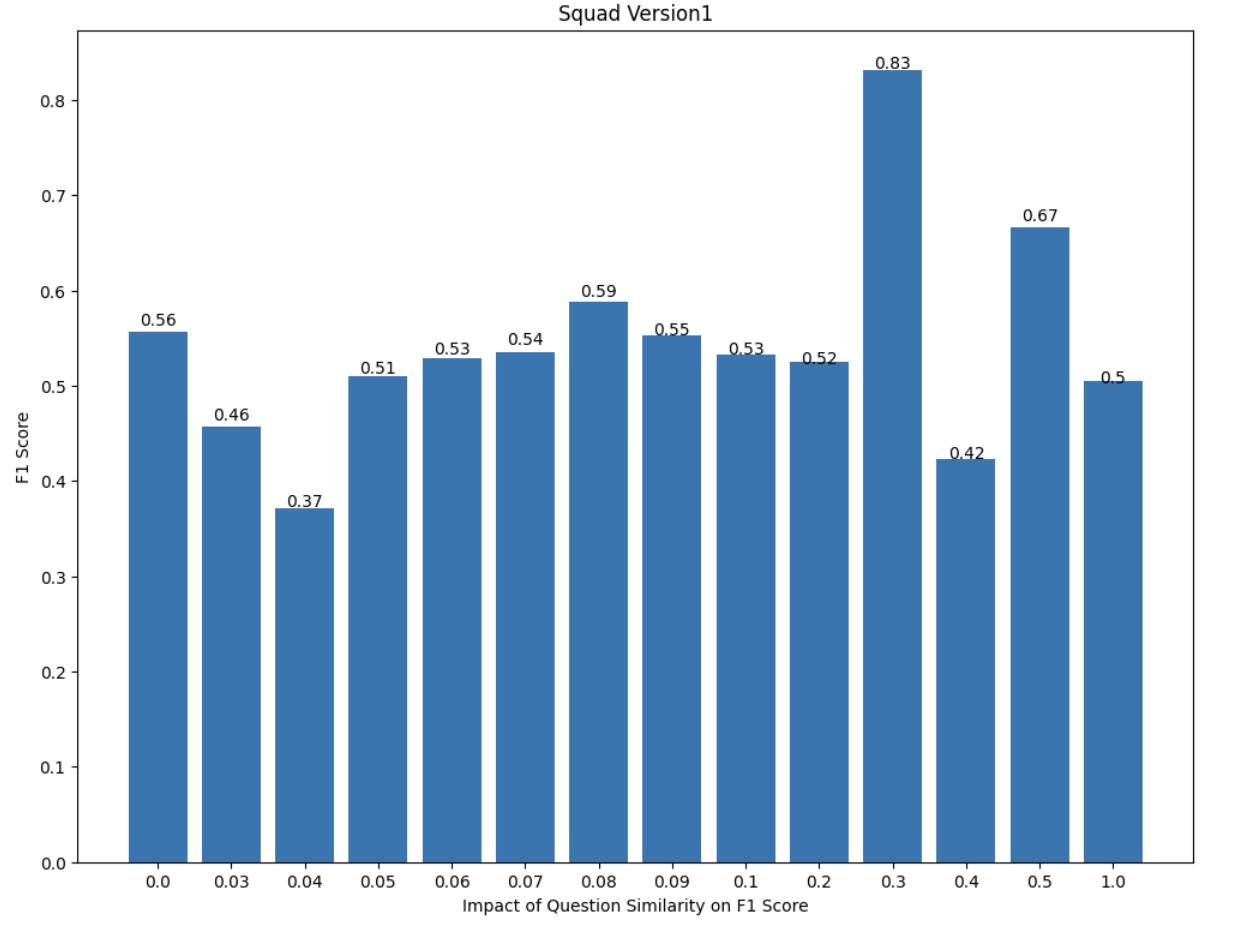}
    \caption{Impact of Question Similarity on F1 Score Performance in SQuAD 1.1}
    \label{fig:qd-squad1}
\end{figure}

\begin{figure}
    \centering
    \includegraphics[width=0.5\linewidth]{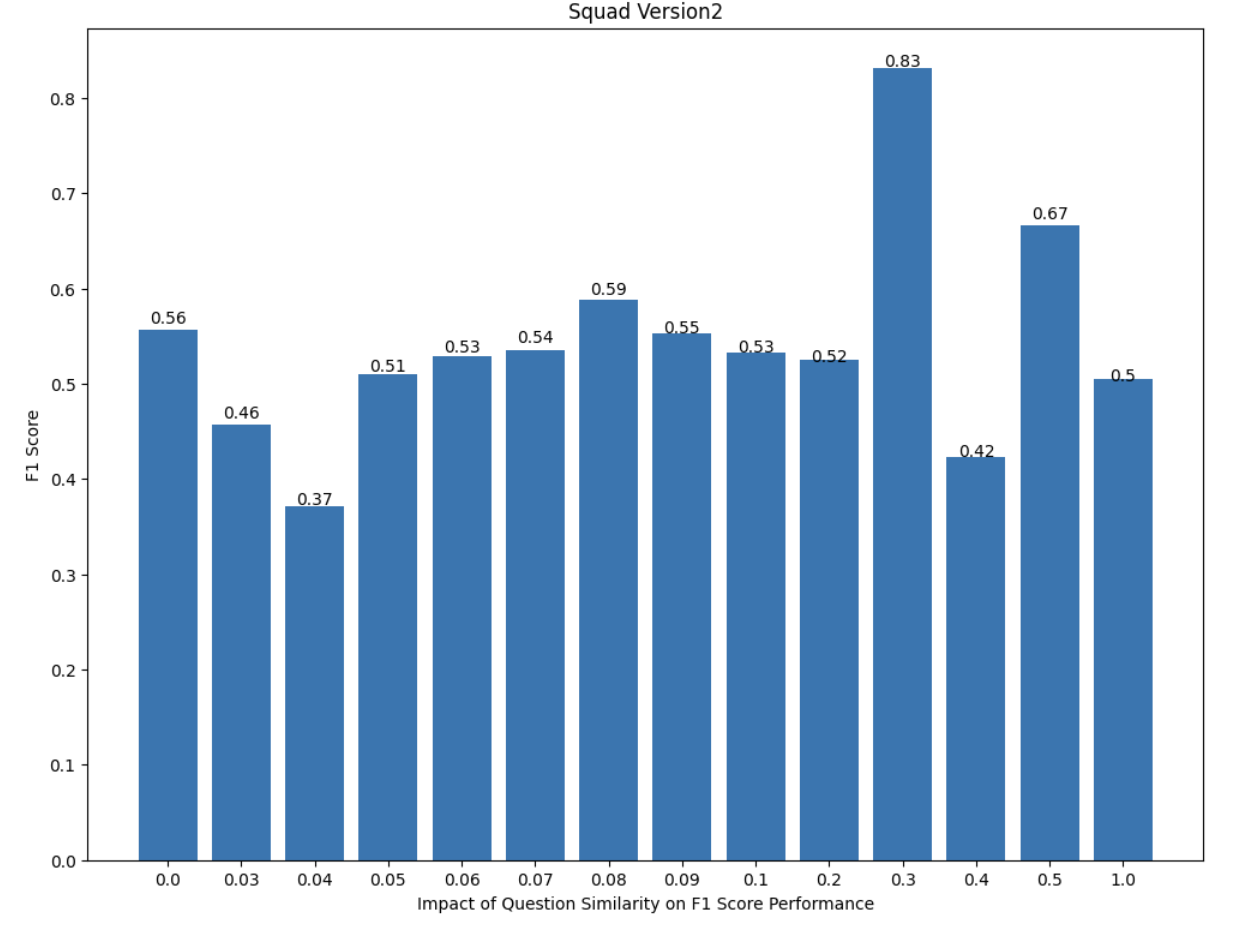}
    \caption{Impact of Question Similarity on F1 Score Performance in SQuAD  2.0 dataset.}
    \label{fig:qd-squad2}
\end{figure}

\begin{figure}
    \centering
    \includegraphics[width=0.5\linewidth]{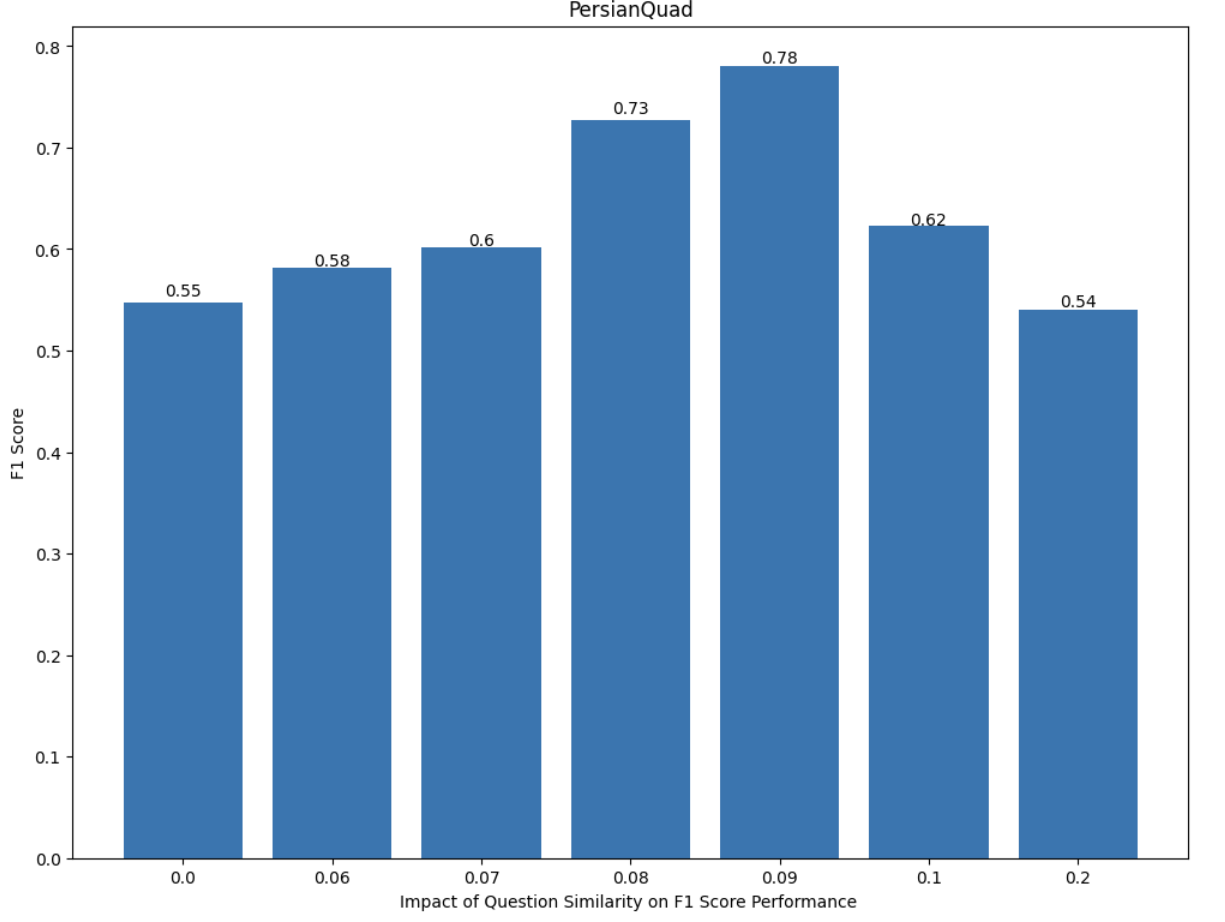}
    \caption{Impact of Question Similarity on F1 Score Performance in PersianQuad.}
    \label{fig:qd-persianquad}
\end{figure}

\begin{figure}
    \centering
    \includegraphics[width=0.5\linewidth]{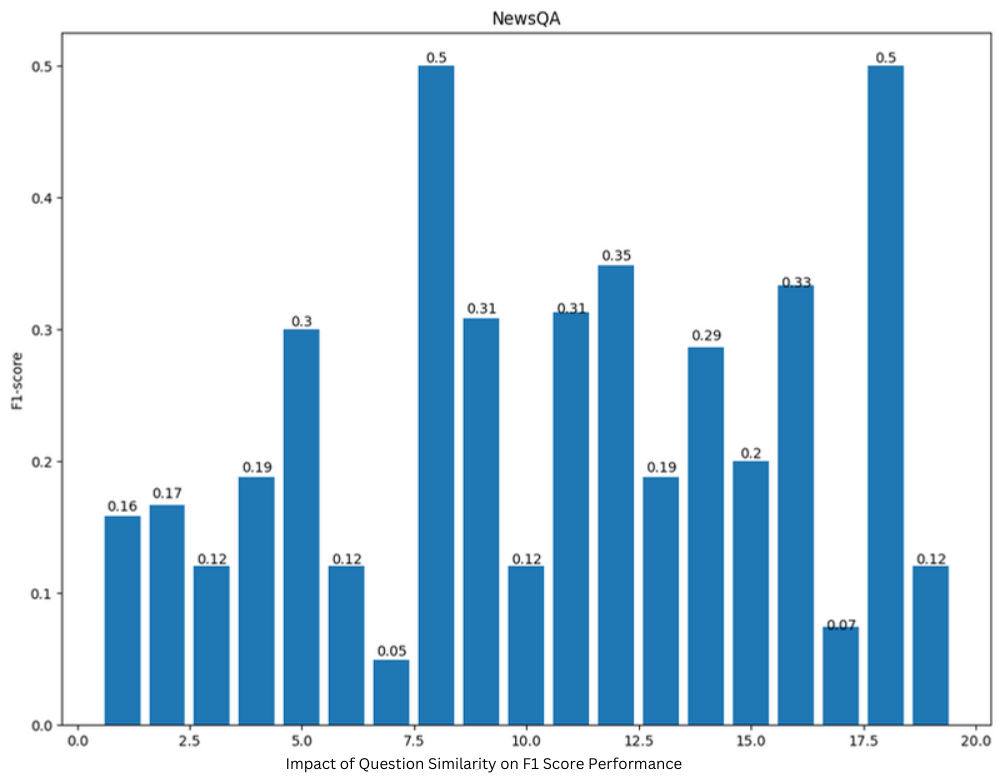}
    \caption{Impact of Question Similarity on F1 Score Performance in NewsQA.}
    \label{fig:qd-newsqa}
\end{figure}

\section{Conclusion}
In this study, we conducted a comprehensive analysis of ChatGPT as a Question Answering System (QAS) and compared its performance with other existing models. Our evaluation focused on the model's ability to extract answers from provided paragraphs, as well as its performance in scenarios without surrounding passages. We explored various aspects, including response hallucination, question complexity, and the impact of context on performance.In conclusion, our comprehensive evaluation of ChatGPT as a QAS has provided valuable insights into its strengths, limitations, and areas for improvement. Leveraging well-known Question Answering (QA) datasets in both English and Persian languages, we employed metrics such as F-score, exact match, and recall to assess ChatGPT's performance. 

Our findings indicate that ChatGPT, while demonstrating competence as a generative model, faces challenges in question answering compared to task-specific models. Context proves to be a crucial factor, with the model exhibiting improved performance when provided with surrounding paragraphs for answer extraction. Prompt engineering, particularly in the form of two-step queries, enhances precision, especially for questions lacking explicit answers in provided passages. We observed ChatGPT's proficiency in answering simpler, factual questions, highlighting its strengths in handling straightforward queries. However, challenges arise with more complex "how" and "why" question types. The evaluation also unveiled instances of hallucinations, where ChatGPT provided responses to questions without available answers in the provided context.

\bibliographystyle{unsrtnat}
\bibliography{template}  






\end{document}